%% file: 00main.tex
\pdfoutput=1

\documentclass[11pt]{article}

\usepackage[final]{acl}

\usepackage{times}
\usepackage{latexsym}

\usepackage[T1]{fontenc}

\usepackage[utf8]{inputenc}

\usepackage{authblk}

\usepackage{microtype}

\usepackage{inconsolata}

\usepackage{multirow}
\usepackage{float}
\usepackage{graphicx}
\usepackage{subfigure}
\usepackage{booktabs}
\usepackage{tabularx}
\usepackage{soul}
\usepackage{enumitem}

\title{
Inquire, Interact, and Integrate: A Proactive Agent Collaborative Framework for Zero-Shot Multimodal Medical Reasoning}

\author[1]{\textbf{Zishan Gu}}
\author[2]{\textbf{Fenglin Liu}}
\author[1]{\textbf{Changchang Yin}}
\author[1]{\textbf{Ping Zhang}}
\affil[1]{The Ohio State University}
\affil[2]{University of Oxford}
\affil[ ]{\{gu.1090, yin.731, zhang.10631\}@osu.edu; fenglin.liu@eng.ox.ac.uk}

\begin{document}
\maketitle
\begin{abstract}
    The adoption of large language models (LLMs) in healthcare has attracted significant research interest. However, their performance in healthcare remains under-investigated and potentially limited, due to i) they lack rich domain-specific knowledge and medical reasoning skills; and ii) most state-of-the-art LLMs are unimodal, text-only models that cannot directly process multimodal inputs. To this end, we propose a multimodal medical collaborative reasoning framework \textbf{MultiMedRes}, which incorporates a learner agent to proactively gain essential information from domain-specific expert models, to solve medical multimodal reasoning problems. Our method includes three steps: i) \textbf{Inquire}: The learner agent first decomposes given complex medical reasoning problems into multiple domain-specific sub-problems; ii) \textbf{Interact}: The agent then interacts with domain-specific expert models by repeating the ``ask-answer'' process to progressively obtain different domain-specific knowledge;  iii) \textbf{Integrate}: The agent finally integrates all the acquired domain-specific knowledge to accurately address the medical reasoning problem. We validate the effectiveness of our method on the task of difference visual question answering for X-ray images. The experiments demonstrate that our zero-shot prediction achieves state-of-the-art performance, and even outperforms the fully supervised methods. Besides, our approach can be incorporated into various LLMs and multimodal LLMs to significantly boost their performance.
\end{abstract}


\input{01introduction}

\input{02related_work}
\input{03methodology}
\input{04experiments}
\input{05conclusion}


\section*{Limitations}
Despite the exceptional performance, our proposed framework still encounters some limitations. Firstly, the performance of MultiMedAgent is constrained by the accuracy of the specialist models. If these domain experts fail to provide the correct knowledge for the learner agent, the conversation could lead more to confusion than clarification. Although we have significantly improved the performance of VQA models through a divide-and-conquer approach, a more robust VQA model would invariably enhance the reliability of our framework. Therefore, developing more advanced VQA models remains a promising direction for future research. Secondly, we observe that sometimes the learner agent delivers all the essential information correctly in its response, yet it receives a lower score than the supervised model simply because it does not adhere to the expected answer format. This issue arises because we are currently employing only natural language processing metrics for evaluation. Incorporating domain-specific evaluation metrics could offer insights from a more practical standpoint. We defer the exploration of such metrics to future studies.

\section*{Ethics Statement}
In this work, we propose an AI agent collaborative framework to provide analysis of medical images in natural text. Our framework enables LLMs, which traditionally lack domain-specific knowledge, to acquire such critical insights regarding medical images from well-trained specialist models and to convey the analysis in the form of conversational dialogue to radiologists. Consequently, our system is not intended to replace radiologists in real-world settings but to act as a supportive assistant, aiding them in clinical decision-making and providing interpretable evidence for the analyses. This approach ensures radiologists can make fully informed decisions independently. We performed our experiments using the public dataset MIMIC-Diff-VQA, which is based on MIMIC-CXR. To comply with the Health Insurance Portability and Accountability Act (HIPAA) standards, all protected health information has been anonymized. Furthermore, all necessary patient/participant consents were obtained, and the appropriate institutional forms have been archived. In adherence to strict privacy protocols, we prohibit the sharing of raw data with any third-party APIs, such as OpenAI and Replicate. For image reference, we use "000A" and "000B" to denote the main image and the reference image, respectively, ensuring that only the VQA answers generated by our local model are shared.

\bibliography{00main}

\appendix
\input{06appendix}

\end{document}

%% file: 01introduction.tex
\section{Introduction}

\label{sec: intro}

\begin{figure}[t]
    \centering
    \includegraphics[width=0.95\linewidth]{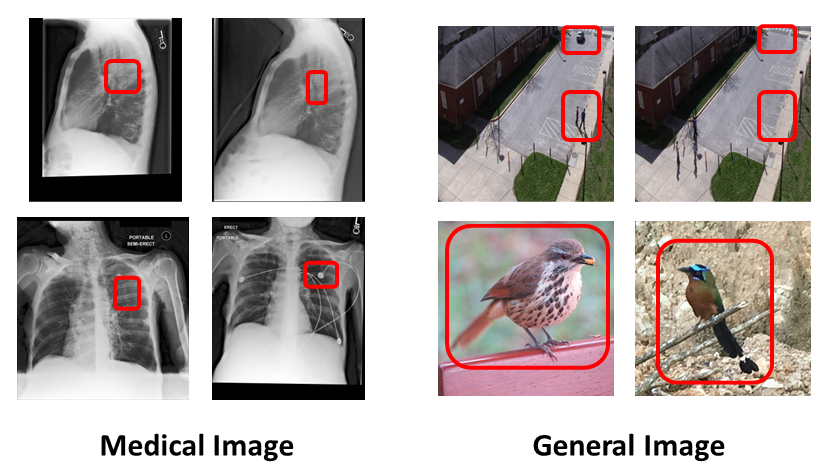}
    \caption{Medical image comparison vs. general image comparison.} 
    \vspace{-4mm}
    \label{fig:introduction} 
\end{figure}

In the domain of medical informatics, the integration of deep learning techniques with extensive hospital database resources has been rapidly evolving these years, particularly in the multimodal analysis of chest X-ray images  \cite{DBLP:conf/cvpr/WangPLLBS17,articlemimic,articlechexpert}. 
This effort has attracted the attention of researchers working on various medical multimodal reasoning tasks, including
the automatic generation of radiological reports \cite{Biswal2020CLARACR,Yin2019AutomaticGO,liu2021exploring,unknownfenglin}, and answering pre-defined medical inquiries \cite{zhang2023pmcvqa,chen2022m3ae}. 
In particular, the task of difference visual question answering (DVQA) in radiology \cite{10.1145/358} addresses the challenge of analyzing sequential images of the same patient taken at different times. This requires the model to answer comparative medical questions that involve assessing changes between images, for example, 'What has changed compared to the previous image?'. Hence, to accurately address these questions, models must not only understand each image effectively and identify abnormalities to assist in diagnosis but also precisely describe the differences between the input images, specifically, the progression of disease. Thus, DVQA in radiology is emphasized as a critical task that mirrors the real-life diagnostic processes closely: clinicians routinely compare sequential X-ray images of the same patient to monitor disease progression.



Answering differences in medical images inherently poses more complexity, especially compared to general images.
As illustrated in \autoref{fig:introduction}, typical DVQA tasks for general images exhibit two characteristics: different images i) may display pronounced differences in their main objects, such as in the case of two distinct birds \cite{forbes2019neural}, or ii) may share the same, usually fixed, viewpoints as those from a video camera, featuring significant changes in salient content \cite{jhamtani-berg-kirkpatrick-2018-learning}. Consequently, clear differences in visual features between images are present, enabling the model to effectively capture their differences.
However, in the context of medical imaging, on the one hand, the main objects (e.g., the abnormalities) remain identical, with sought-after differences residing in subtle details (e.g., the severity and the size of the abnormalities). 
Moreover, it is the identical regions that predominantly feature across images, rather than the differing ones. As shown in \autoref{fig:introduction}, the notable differences, highlighted by red bounding boxes, constitute only a minor portion of the overall images. This leads to the visual differences between medical images being relatively understated, thereby hindering the model's ability to discern differences through direct feature comparison. Instead, a model must employ a higher, more abstract level of understanding and comparison to accurately identify changes \cite{10.1145/358}. 
On the other hand, X-ray images of the same patient, even when taken in the same body position, can vary in viewpoints and scale, further making discrimination more difficult. Therefore, to answer differences between X-ray images, a model necessitates a comprehensive understanding of varied domain-specific knowledge to accurately identify a wide range of clinical findings, which may present more significant differences than those discernible through visual features alone. 
Furthermore, the model must also be capable of recognizing the subtle yet critical differences within the main objects (e.g., disease progress), which are pivotal for effective disease monitoring.


\begin{figure}
  \centering
  \includegraphics[width=\linewidth]{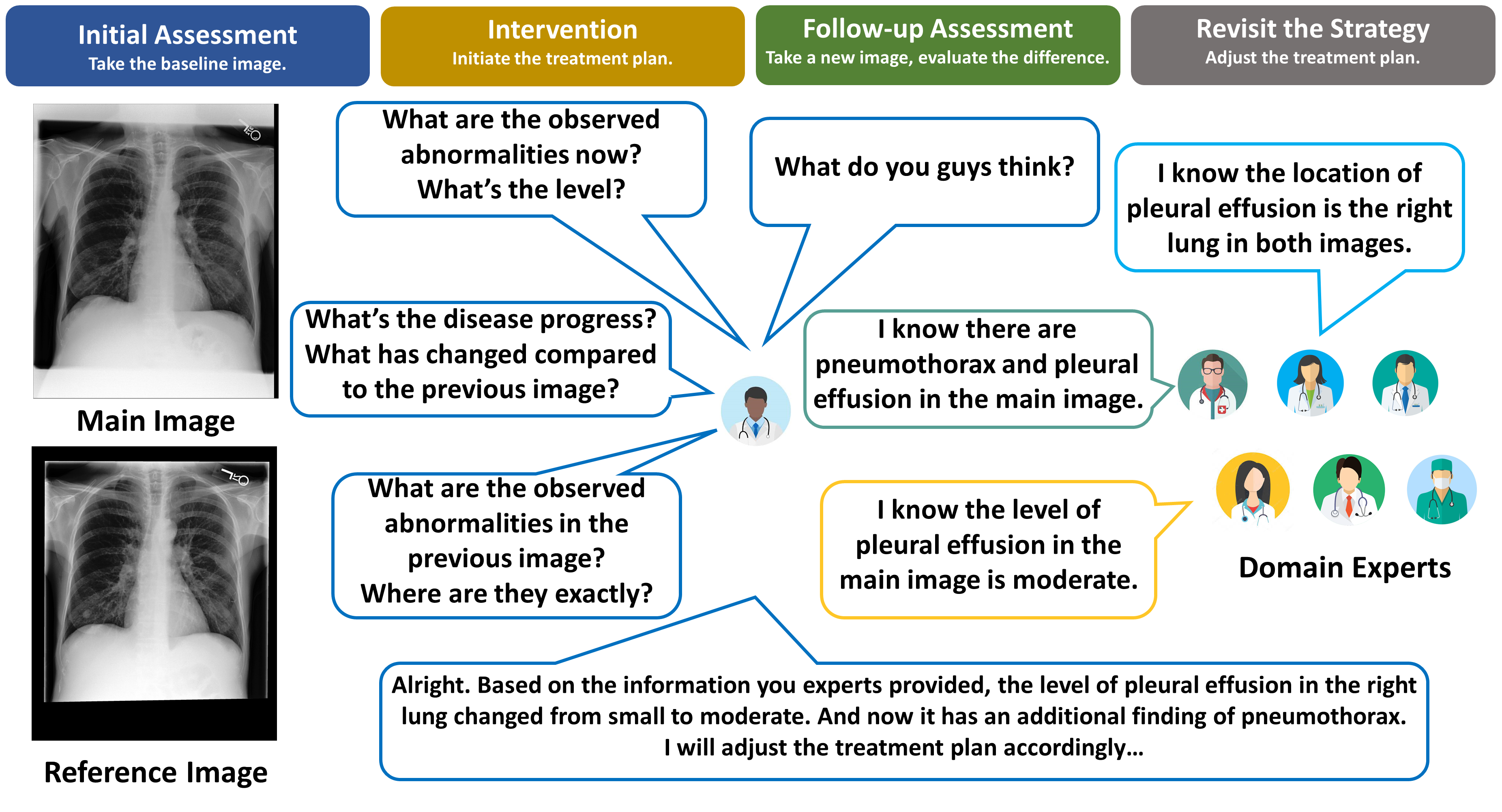}
  \caption{Illustration of the expert consultation. 
  }
  \label{fig: 0}
  \vspace{-4mm}
\end{figure}

To effectively capture domain-specific knowledge and the subtle yet critical differences between chest X-ray images, we propose a multimodal medical collaborative reasoning framework (MultiMedRes), which imitates the clinicians’ working patterns.
As shown in \autoref{fig: 0}, the common practice of a patient treatment process \cite{delrue2011difficulties}
typically commences with the acquisition of a baseline image to serve as diagnostic evidence and the foundation for initiating a treatment plan. Subsequently, a follow-up image is obtained to monitor and assess the intervention's efficacy throughout the treatment period. This process often involves consultations with multiple experts across various domains, which may lead to modifications of the initial treatment plan.
To model the above working patterns, MultiMedRes introduces three steps led by a learner agent:
i) \textbf{Inquire}: The learner agent aims to understand the broad 
challenging difference questions and ask multiple types of domain-specific sub-questions that target specific aspects of the images, e.g., detecting abnormalities, identifying the level of severity and the location of the disease; 
ii) \textbf{Interact}: The learner agent feeds the sub-questions into expert models (i.e., specialists) to obtain their answers.
These expert models are pre-trained on the domain-specific task and data to store rich specific knowledge for each type of domain-specific question.
Our method will then raise new questions based on the given answers. Through repeating the ``ask-answer'' interaction process, our agent can progressively obtain sufficient knowledge for different domain-specific questions from the expert models;
and iii) \textbf{Integrate}: The learner agent finally integrates all knowledge from the domain-specific specialists to address the input difference question accurately.
These three steps can also provide readily interpretable for human radiologists.

We conduct the experiments on the benchmark MIMIC-Diff-VQA \cite{10.1145/358} dataset. Experiments demonstrate that our approach can achieve state-of-the-art performances.
Moreover, MultiMedRes can be incorporated into diverse LLMs and multimodal LLMs to significantly
boost their performance. The contributions of our study are outlined as follows:

\begin{itemize}[leftmargin=*]
    \item We introduce a collaborative reasoning framework, MultiMedRes, which
    enables an LLM learner agent to acquire essential domain-specific knowledge from specialized expert models to perform zero-shot medical multimodal reasoning.
    
    \item 
    The zero-shot prediction provided by our framework achieves state-of-the-art performance, and even outperforms the fully supervised methods. 
    
    \item  Experiments demonstrate that our framework is compatible with various LLMs, including both text-based LLMs and multimodal LLMs.
\end{itemize}

%% file: 02related_work.tex
\section{Related Work}
\label{sec: related work}

Currently, multi-modal LLMs have also demonstrated promising results in tasks related to medical imaging \cite{lee2023llmcxr,xu2023elixr,Thawakar2023XrayGPTCR}, including multimodal reasoning.
There have been explorations leveraging the extensive knowledge base and advanced reasoning capabilities of large language models (LLM) \cite{lu2022unifiedio,gui-etal-2022-kat,DBLP:journals/corr/abs-2109-05014}. 
While these large-scale models indeed shed light on tackling multiple multimodal tasks in the medical domain all at once, they have not surpassed traditional state-of-the-art models for multimodal reasoning on widely recognized datasets \cite{chen2024chexagent}. 
Besides, such large-scale models all necessitate extensive training or fine-tuning of enormous amounts of parameters and are not tailored for low-resource settings (e.g., rare diseases and novel diseases), where the labeled data for training is limited.
In this work, we propose a collaborative agent framework to enable the LLMs to accurately perform zero-shot multimodal reasoning. 

Recently, with the growing research interest in LLMs, the concept of LLM-driven collaborative multi-agent systems has garnered significant attention \cite{anonymous2023adapting,park2023generative}. 
LLM agents have also shown considerable promise in managing complex tasks in medical imaging \cite{allioui2021optimized,BENNAI2020101980,jpm12020309}. 
Nevertheless, to the best of our knowledge, research on leveraging this paradigm for medical multimodal reasoning remains sparse.

%% file: 03methodology.tex
\section{Methods}
\begin{figure*}[h]
    \centering
    \includegraphics[width=0.95\textwidth]{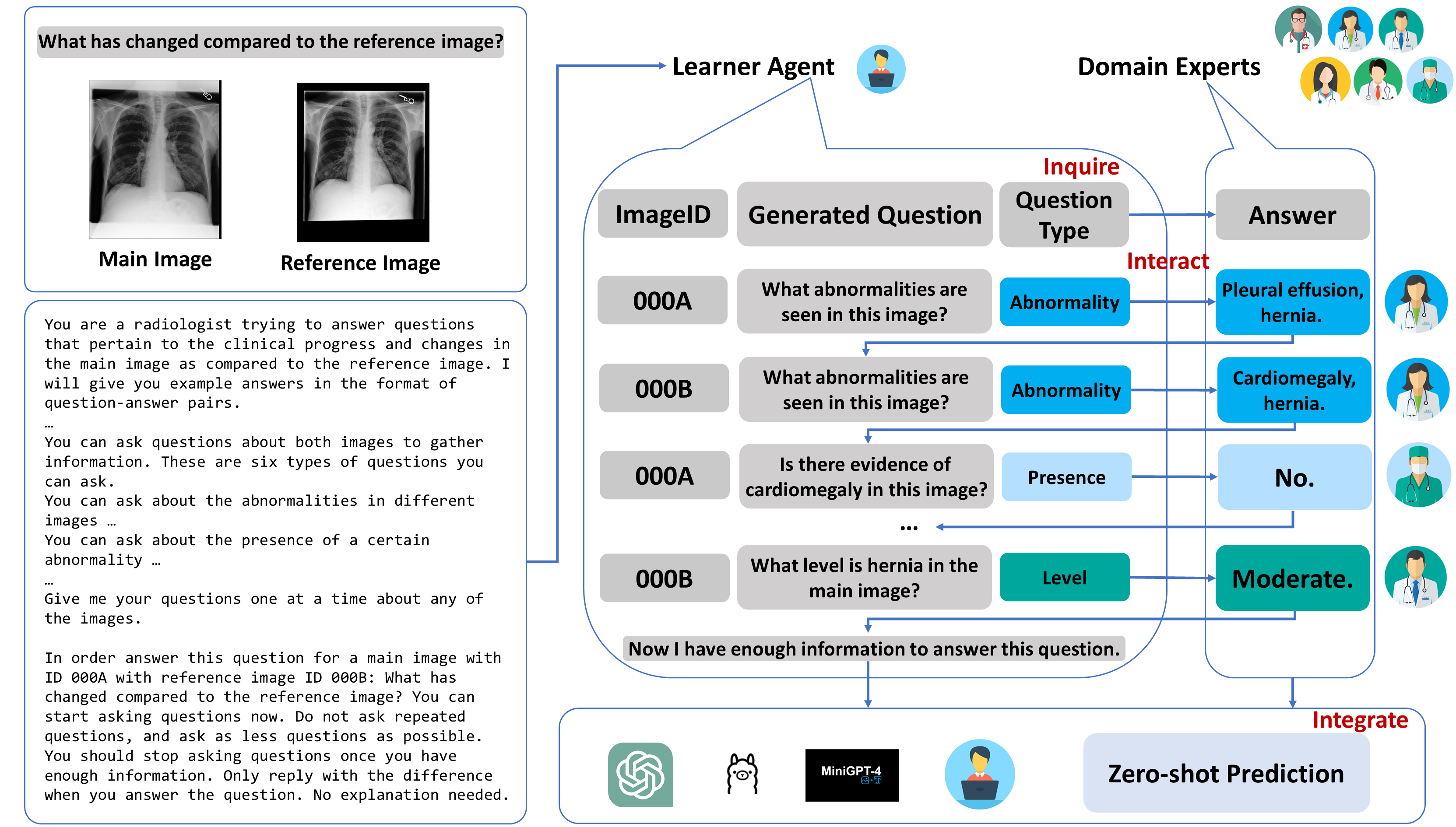}
    \vspace{-3mm}
    \caption{The proposed MultiMedRes framework. 
    Upon receiving questions comparing two images, the learner agent employs an iterative approach, generating questions related to either the main image or the reference image, before consulting the appropriate domain expert (i.e., specialists). Upon collecting sufficient information, the learner agent is prompted to cease question generation and integrate the information to provide a zero-shot prediction. 
    } 
    \label{fig: 1} 
    \vspace{-4mm}
\end{figure*}
In this section, we introduce MultiMedRes, a novel collaborative framework for medical reasoning tasks. 
Initially, we train a cohort of domain expert models, each tailored to address a distinct category of questions (e.g., abnormality detection and localization), to serve as domain-specific experts. Subsequently, an LLM learner agent is designed to generate inquiries and interact with these experts to gather the essential information for the medical reasoning questions.
Finally, once the agent obtains enough information, it will stop asking questions and integrate the conversation for answering. 
\autoref{fig: 1} displays the overview of the proposed MultiMedRes.


\subsection{Problem Statement} 

For common medical questions involving a single image (VQA), we adhere to the standard practice of VQA tasks by implementing classification models. Our trained domain expert models, denoted as $\mathcal{F}_{sp}$, are designed to receive an image $i\in I$ and a question $q\in Q$ as inputs, and to subsequently predict the label $a\in A$, which corresponds to answers observed in the dataset. Further, in this study, we mainly focus on the challenging yet essential medical reasoning task: difference question answering (i.e., DVQA). This type of questions inquires about the comparison between current and previous images of the same patients and studies the treatment or disease progress, which is more consistent with real-life radiologists’ practice. As illustrated in \autoref{fig: 1}, our framework would take two images $i_1, i_2\in I$ and a difference question $q_d \in Q$ as input and output the difference captioning regarding the question. More specifically, the agent $\mathcal{F}_{learner}$ will generate questions concerning single images iteratively and interact with the corresponding specialist models to derive an answer in each round of the conversation, terminating the process upon acquiring sufficient information to formulate a response to the initial difference question.

In the subsequent section, we aim to design our prompting system to assist the LLM agent $\mathcal{F}_{learner}$ in comprehending the task, formulating the appropriate questions, and concluding the conversation appropriately.


\subsection{Learner Agent}
In MultiMedRes, we introduce an automatic question generation mechanism that leverages LLM’s in-context learning and reasoning abilities. We utilize the LLM as a learner agent to generate informative questions as well as the corresponding question types about chest X-ray images, and direct these questions to well-trained domain-expert models for answer retrieval. 

Mimicking the reasoning process of a human being when answering a difference question, a learner agent must first gather information of the main image and the reference image individually, and then proceed to summarize the difference. 
Specifically, we incorporate ChatGPT \cite{NEURIPS2020_1457c0d6} or LLaMa \cite{touvron2023llama}  as the learner agent responsible for generating questions.
To optimize the in-context learning and reasoning abilities of the LLMs,
we design the prompting system with three parts: a systematic instruction $\rho_{task}$ to describe the difference question-answering task, a context-learning instruction $\rho_Q$ to guide question generation, and an appended instruction $\rho_i$ to signal the end of the questioning process and initiate summarization. Note that during the conversations, the generated question-answer pairs in earlier rounds (denoted as $\rho_{log}$) are also visible to the agent. Consequently, every question is generated using the combined prompt of $\rho_{task} + \rho_Q + \rho_i + \rho_{log}$. 

\subsubsection{Task Instruction $\rho_{task}$}

The task instruction $\rho_{task}$ outlines the task that the learner is required to perform and provides the material for in-context learning of the difference question answering. It guides the LLM to utilize the related knowledge base and generate questions to gather necessary information on images to compare the differences. $\rho_{task}$ is designed as follows:\vspace{2mm}

\textit{You are a radiologist trying to answer questions that pertain to the clinical progress and changes in the main image as compared to the reference image. I will give you example answers in the format of question-answer pairs. [context]
Please answer this question for a main image $A$ with reference image $B$: [difference questions] } 
\vspace{2mm}

The initial sentence directs the learner agent to access and apply radiology-related knowledge bases, while the context will demonstrate the desired answer formats (as shown in \autoref{sec: prompt}). 

\subsubsection{Question Generation Instruction $\rho_Q$}
To assist the learner agent in question generation, we supply both the categories and contents of the questions, which facilitates the identification of the most appropriate domain expert model. $\rho_Q$ is structured as follows:
\vspace{2mm}

\textit{You can ask questions about both images to gather information, but do not ask redundant questions. 
You can ask about the abnormalities in different images, like [...]. 
You can ask about the presence of a certain abnormality in an image, like [...]. 
You can ask about the level of a certain abnormality in an image with [...].
Give me your questions one at a time about any of the images. Only return the generated question, the question type, and the corresponding image ID.}
\vspace{2mm}

[...] above represents the corresponding question format of the given types, which will guide the LLM to generate the desired questions.

\subsubsection{Appended Instruction $\rho_i$}
From an efficiency perspective, the agent should stop asking questions once it gathers enough information, and answer with simplicity. Thus, $\rho_i$ as the last part of the prompt is structured as follows:
\vspace{2mm}

\textit{You should answer the question like the previous examples once you have enough information. Do not make any assumptions by yourself. Only reply with the difference when you answer the question. No explanation is needed.} \vspace{2mm}

Please refer to the \autoref{sec: prompt} for a full version of the finalized prompt. 


\subsection{Domain Expert Models}

As discussed in \autoref{sec: related work}, despite the significant efforts towards developing medical vision LLMs recently, the performance of these models on specialized medical VQA datasets has failed to surpass that of the traditional state-of-the-art classification models. 
Consequently, in this work, we continue to rely on classification models as the domain experts. However, take MIMIC-DIFF-VQA \cite{10.1145/358} dataset shown in \autoref{tab: static} in \autoref{sec: statistics} as an example, there are totally over $9000$ answer candidates for all types of questions, not including those for difference questions. This vast number of labels could potentially overwhelm a classification model, particularly when the candidates are only marginally distinct from one another. 
Moreover, the amount of potential answers for abnormality-related questions is enormous as over $8000$. This is particularly evident in responses to the question, "What abnormalities are seen in this image?", where answers comprise various combinations and permutations of 33 abnormalities, further complicating the differentiation process among these answers. For example, the answer "atelectasis, penumothorax" and the answer "penumothorax, atelectasis" are the same thing, and they are only slightly different than "penumothorax, atelectasis and pleural effusion". 

To mitigate this challenge, our framework employs a divide-and-conquer strategy. Specifically, we develop a distinct expert model for each question type. As in MIMIC-DIFF-VQA, we will train an expert model for the "Abnormality", "Presence", "View", "Location", "Type" and "Level" type of questions separately as shown in the $2^{th}-7^{th}$ rows in \autoref{tab: static}. This approach significantly narrows the scope of potential answers, enabling these specialists to concentrate exclusively on their respective domains. Moreover, regarding the 'abnormality' question, which essentially involves detecting abnormalities throughout the entire image, we shift our strategy. Rather than trying to match answers from an extensive training set containing a huge amount of responses, we focus on creating an abnormality detection model tailored specifically to this task, akin to multi-label classification. This model processes the image input and generates a list of identified abnormalities, as exemplified in the chest X-ray images. The response to this question is then formed by simply concatenating the predicted labels. Consequently, in MIMIC-DIFF-VQA, the number of possible answers for subsequent 'abnormality' questions is reduced to $25$, a volume well-suited for a classification model.

\subsection{Agent Behavior}

As illustrated in Figure~\ref{fig: 1}, for questions involving the comparison of two images, or difference questions, our proposed MultiMedRes follows an iterative pattern. During the initial iteration, the learner agent formulates a query for the expert models based on the question only. For the following iterations, the agent not only considers the difference questions but also integrates information already acquired from earlier conversations. This strategy enables the LLM agent to generate the next question accordingly and determine the appropriate time to cease inquiries, thereby avoiding the generation of redundant or irrelevant questions.

%% file: 04experiments.tex
\section{Experiment}

\begin{table*}[!htbp]
    \footnotesize
    \centering
    \begin{tabular}{lccccccc}
    \toprule
    Models                              & Bleu-1 & Bleu-2 & Bleu-3 & Bleu-4 & METEOR & ROUGE\_L & CIDEr \\ 
    \midrule
    EKAID \cite{10.1145/358}            & 0.569  & 0.498  & 0.438  & 0.382  & 0.304  & 0.547    & 0.823 \\ 
    EKAID\_diff  \cite{10.1145/358}                         & 0.606  & 0.529  & 0.468  & 0.410  & 0.350  & 0.572    & 0.827 \\ 
    UIO \cite{lu2022unifiedio}          & 0.360  & 0.309  & 0.267  & 0.223  & 0.220  & 0.426    & 0.388 \\
    MiniGPT-v2 \cite{chen2023minigptv2} & 0.291  & 0.237  & 0.190  & 0.146  & 0.333  & 0.391    & 0.110 \\ 
    LLaVa \cite{liu2023llava}           & 0.411  & 0.333  & 0.257  & 0.185  & 0.320  & 0.452    & 0.162 \\
    Med-Flamingo \cite{moor2023medflamingo}  & 0.573  & 0.505  & 0.433  & 0.359  & 0.304  & 0.544    & 0.438 \\ 
    \midrule
    MultiMedRes (GPT-3.5)                & 0.497  & 0.423  & 0.361  & 0.303  & 0.317  & 0.526    & 0.451 \\ 
    MultiMedRes (LLaMa2)                 & 0.537  & 0.465  & 0.391  & 0.345  & 0.373  & 0.554    & 0.483 \\ 
    MultiMedRes (GPT-4.5)                & \textbf{0.610}  & \textbf{0.535}  & \textbf{0.473}  & \textbf{0.418}  & \textbf{0.357}  & \textbf{0.586}    & \textbf{0.843} \\ 
    \bottomrule
    \end{tabular}
    \caption{Comparative performance of various models on the difference question answering task. EKAID and EKAID\_diff represent the state-of-the-art task-specific models, which adopt the labeled data to perform fully-supervised training.
    The last three rows represent the zero-shot performance of MultiMedRes with the model being GPT 3.5, LLaMa2, and GPT 4.5 respectively. 
    } 
    \label{tab: main results}
\end{table*}

\subsection{Experimental Setting}
\paragraph{Dataset and Setup}
In this study, we conduct comprehensive experiments on the newly-released DVQA dataset of chest x-ray images, MIMIC-Diff-VQA\cite{10.1145/358}, which was derived from the open benchmark dataset MIMIC-CXR \cite{articlemimic}. It includes a total of seven types of questions, including $164,324$ questions that require the comparison of two images, referred to as difference questions. We provide a more detailed description of the dataset in \autoref{sec: statistics}.

We employ both GPT \cite{NEURIPS2020_1457c0d6} accessed through the OPENAI API \footnote{https://openai.com/} and LLaMa 2 \cite{touvron2023llama} via the Replicate API \footnote{https://replicate.com/} as the learner agents. For the domain expert models, we integrate MMQ \cite{aioz_mmq_miccai21} for the VQA tasks and DenseNet \cite{Huang2016DenselyCC} for abnormality detection. Due to the space restriction, implementation details of each integrated model are given in \autoref{sec: implementation}. As for the evaluation, following previous work, we use the popular natural language processing metrics BLEU \cite{10.3115/1073083.1073135}, METEOR \cite{10.5555/1626355.1626389}, ROUGE\_L \cite{lin-2004-rouge}, CIDEr \cite{7299087} for the performance on difference question answering. For the zero-shot prediction provided by the LLMs, we report the performance of a single run due to budget constraints. For other question types, we adhered to the common practice of using accuracy as the evaluation metric. 

\paragraph{Baselines}
In this work, we consider a state-of-the-are medical VQA model, MMQ \cite{aioz_mmq_miccai21}, a state-of-the-art medical DVQA model, EKAID \cite{10.1145/358} and three multi-modal LLMs, UNIFIED-IO\cite{lu2022unifiedio},  MiniGPT-v2 \cite{chen2023minigptv2} and LLaVA \cite{liu2023llava} as baselines. Please refer to \autoref{sec: baseline} for details.

\subsection{Main Results}

\begin{figure}[t]
\centering
    \subfigure[MiniGPT]{\includegraphics[width=1.4in]{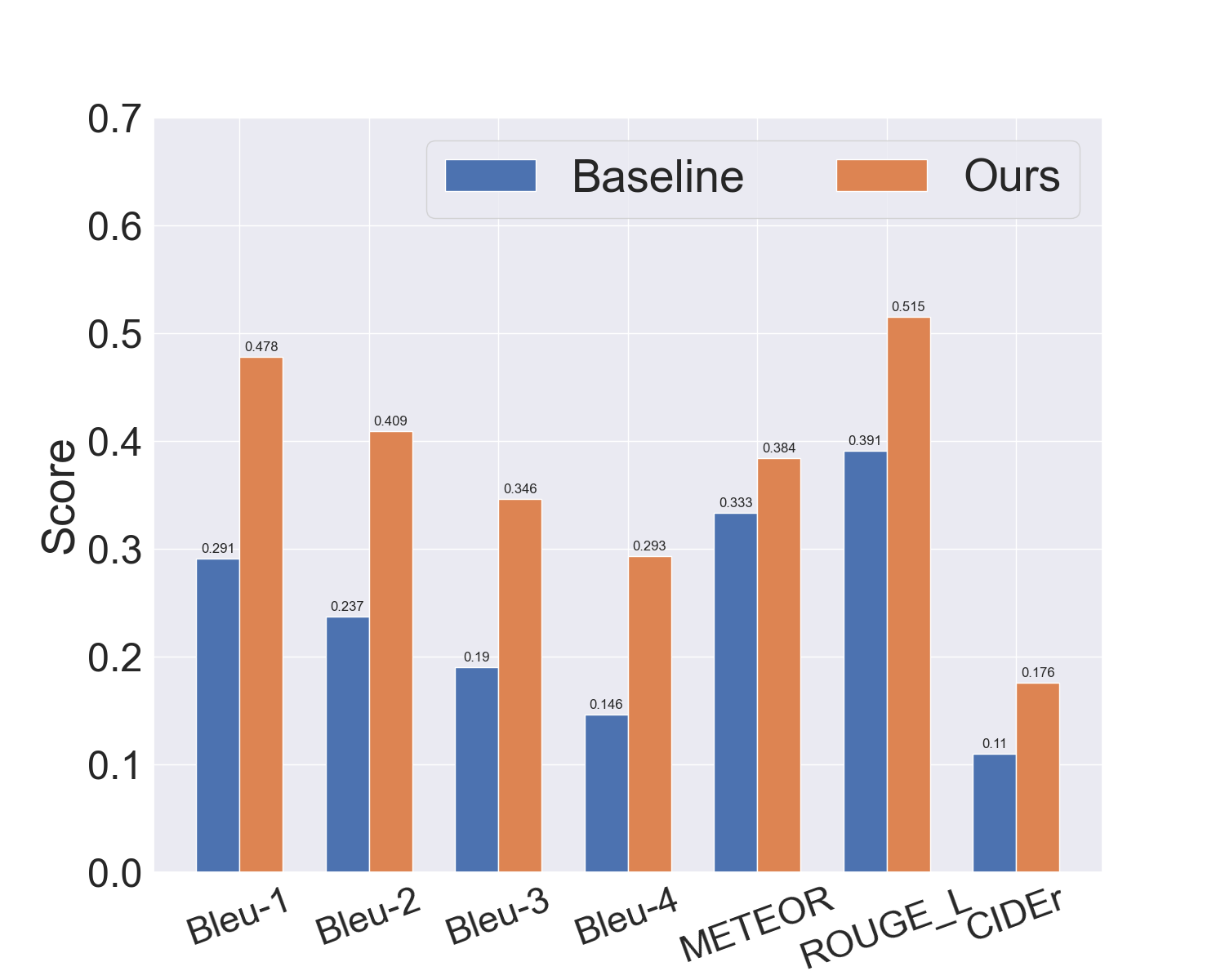}\label{fig: MiniGPT}}
    \subfigure[LLaVa]{\includegraphics[width=1.4in]{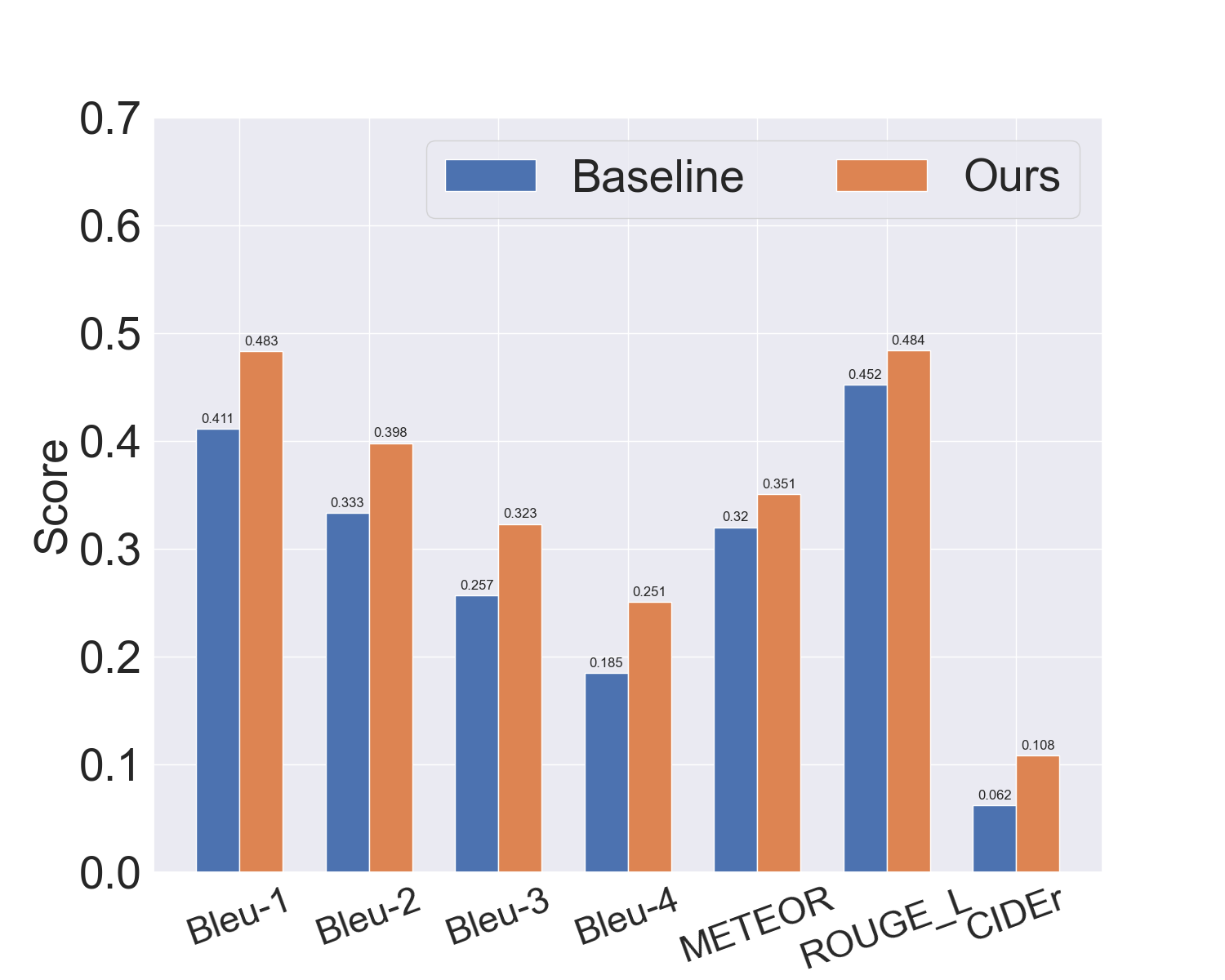}\label{fig: LLaVa}}
    \vspace{-3mm}
    \caption{Performance comparison for vision LLMs with or without our MultiMedRes. As we can see, our method significantly boosts the performance of vision LLMs across all metrics.}
    \vspace{-4mm}
    \label{fig: context}
\end{figure}

We report the models' performance with difference questions in \autoref{tab: main results}, and the performance of expert models on questions regarding single images in \autoref{tab: VQA results} in \autoref{sec: VQA performance}. It can be observed that, training the EKAID model solely with difference questions results in a moderate improvement in performance compared to the baseline established by the original work. As for the large-scale multimodal models, specifically Unified-IO\cite{lu2022unifiedio}, MiniGPT-v2\cite{chen2023minigptv2}, and LLaVa\cite{liu2023llava}, they all struggle to attain competitive performance on both DQA task and VQA task, likely attributable to a deficiency in domain knowledge. Conversely, leveraging the significantly enhanced capabilities of domain expert models, facilitated by our divide-and-conquer strategy, the zero-shot predictions generated by GPT-3.5, LLaMa2-70B, and GPT-4-Turbo through our proposed MultiMedRes all demonstrate competitive performance. Notably, GPT-4-Turbo significantly surpasses the previous state-of-the-art models, including the fine-tuned domain-specific LLM and the fully-supervised generative model, without requiring specific training on DVQA data. Furthermore, as demonstrated in \autoref{fig: context}, incorporating the dialogue between the learner agent and the specialists into the prompts for the vision LLMs, e.g. MiniGPT-v2 and LLaVa, markedly enhances their performance. This improvement, combined with the outstanding results from the zero-shot prediction capability of MultiMedRes, suggests that the dialogue between the learner agent and the specialists indeed encapsulates essential domain knowledge absent in the LLMs, shedding light on the adaptive usage of LLMs in specific domains.

\begin{figure}[t]
\centering
    \vspace{-3mm}
    \subfigure[Bleu-4]{\includegraphics[width=1.4in]{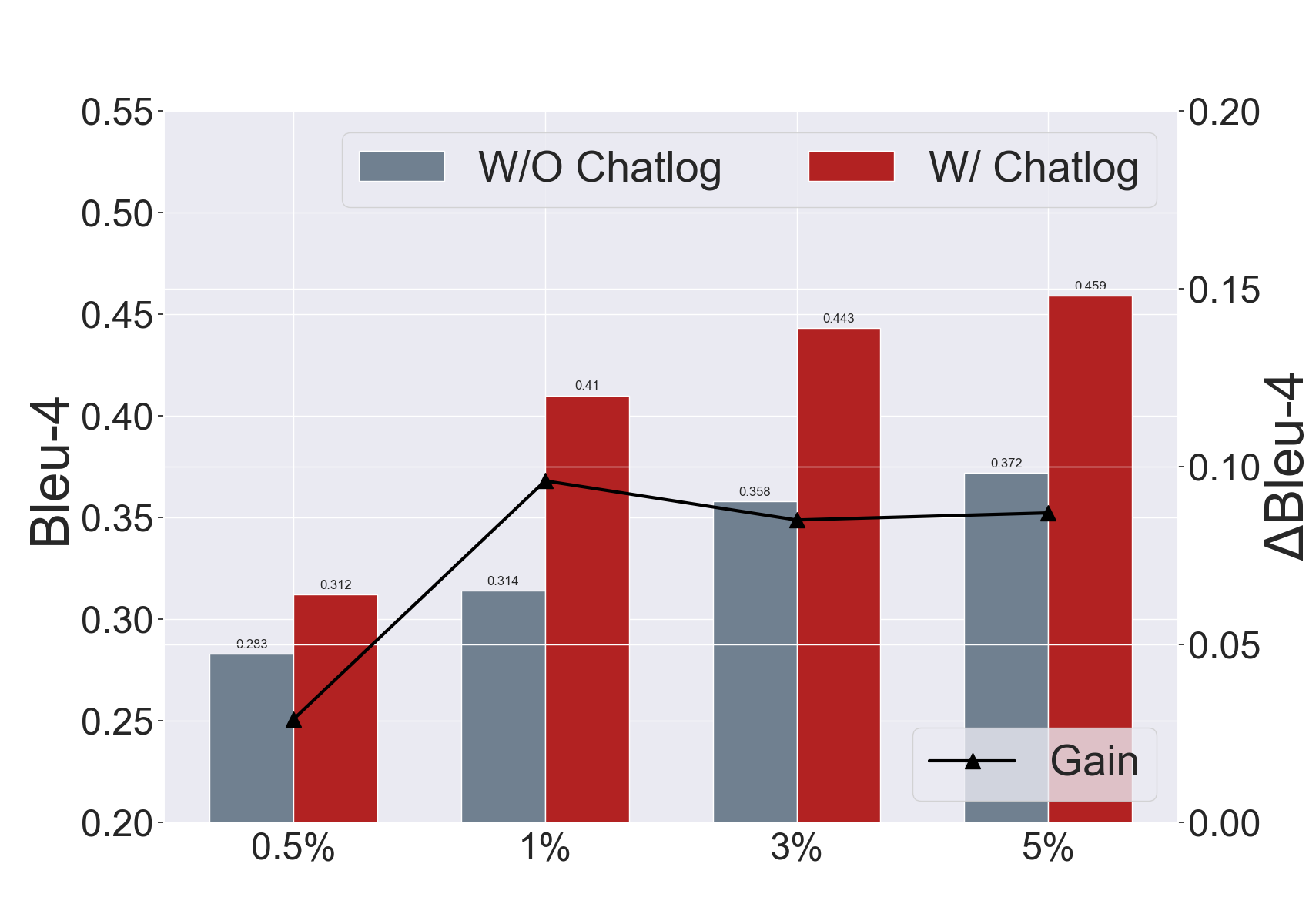}\label{fig: Bleu}}\ \ \
    \subfigure[ROUGE\_L]{\includegraphics[width=1.4in]{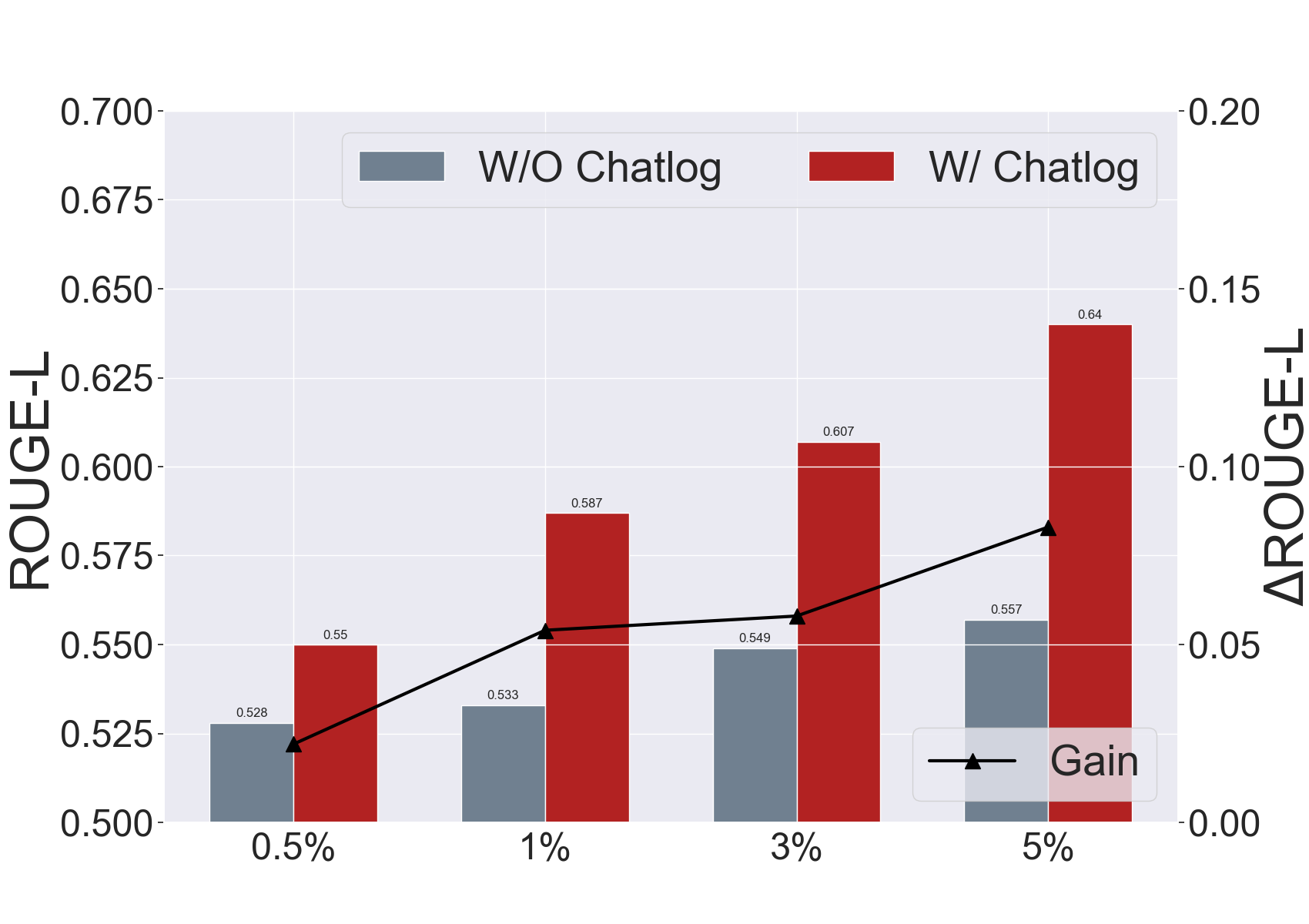}\label{fig: ROUGE-L}}\ \ \
    \vspace{-4mm}
    \subfigure[METEOR]{\includegraphics[width=1.4in]{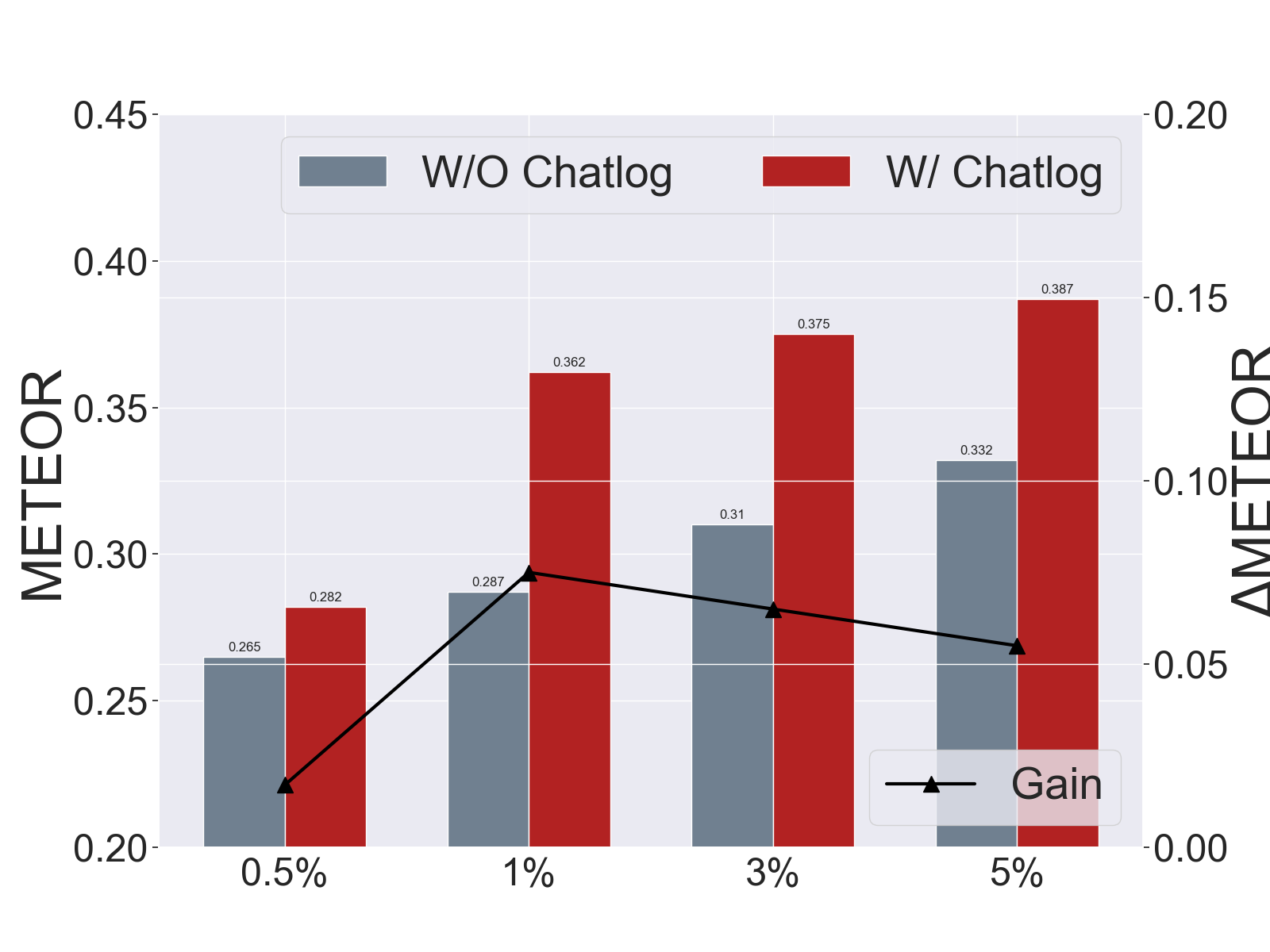}\label{fig: METEOR}}\ \ \
    \subfigure[CIDEr]{\includegraphics[width=1.4in]{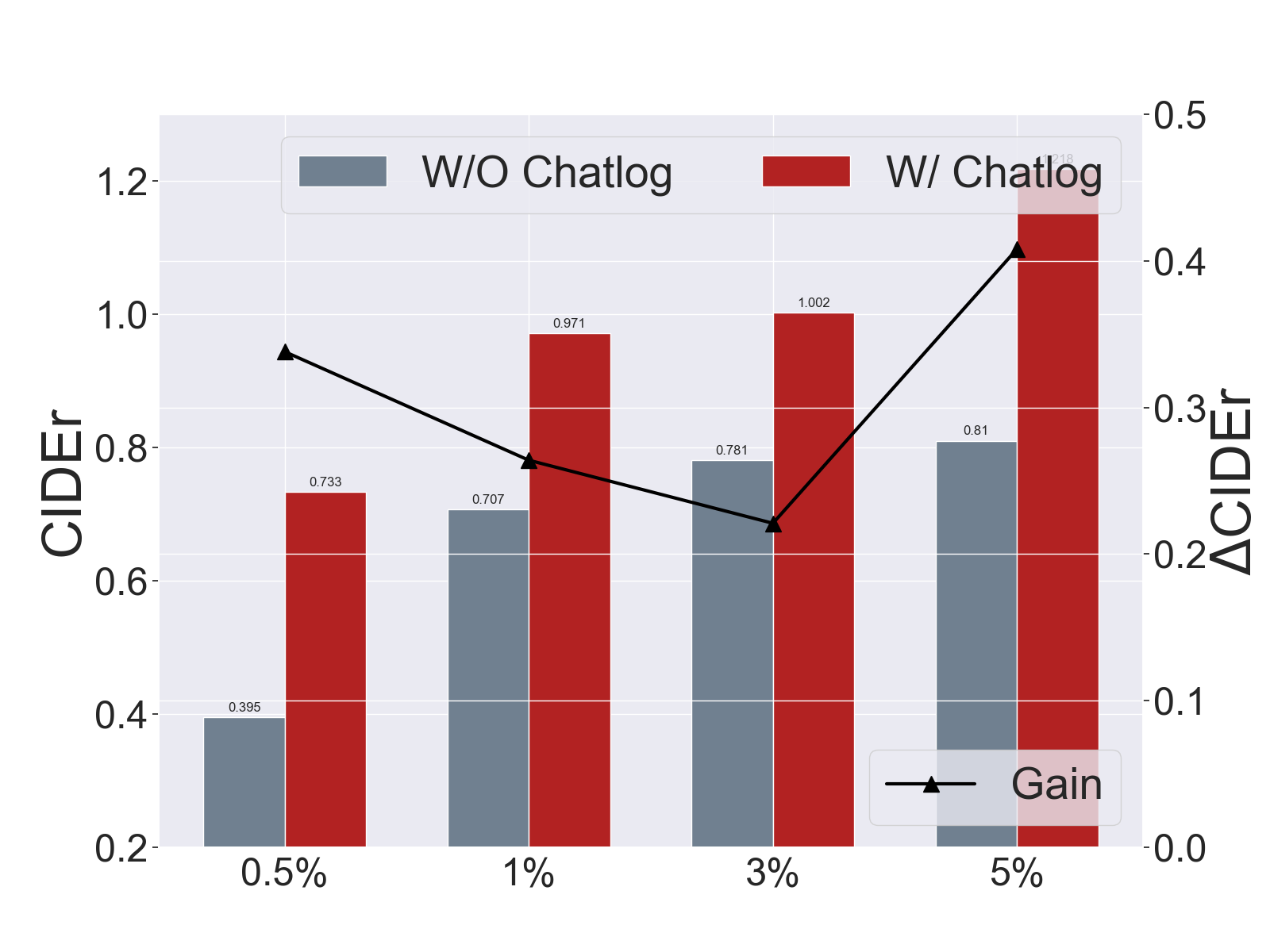}\label{fig: CIDEr}}\ \ \
    \caption{Bleu-4, METEOR, ROUGE\_L and CIDEr score of the few-shot prediction generated by EKAID with respect to various ratios of labeled data (i.e., question-answer pairs) for training. The comparison is between the model trained with or without the augmented training data. The differences at different ratios are depicted using a polyline and the scales are indicated on the right y-axis.} 
    \vspace{-6.5mm}
    \label{fig: few-shot}
\end{figure}

\section{Analysis}
\subsection{Augmented Training Data}
To demonstrate the generalizability of our proposed framework, we further incorporate our generated conversation between the learner agent and the specialists as part of the training data for the supervised model EKAID. Specifically, instead of only considering the difference question as the text input, we also input the conversation chatlog for the training and test samples. We present the performance of EKAID with varying proportions of the training set in \autoref{fig: context}. It is observed that our generated chatlog enables the model to consistently outperform one trained without it, and it further improves its performance with an increase in training data. Notably, with only $1\%$ of training data, the EKAID model with augmented training data achieves comparable performance with the one trained with all of the training data without chatlog. With $5\%$ of the training data, it also outperforms the zero-shot prediction generated by the MultiMedRes. This is because the enhancement of zero-shot learning benefits from, and at the same time, is limited by the answers provided by the domain expert models. In other words, its performance is largely determined by the answer accuracy of questions regarding single images. On the other hand, while LLMs tend to accept whatever the specialists return and include everything in their answers, the supervised few-shot prediction model with augmented training data has the opportunity to identify and correct the incorrect, retaining only the essential information in the final answer. We will further discuss this difference in the case study in \autoref{sec: case study}.

\subsection{Conversation Study}
\label{sec: case study}
\begin{figure}[h]
    \centering
    \includegraphics[width=0.45\textwidth]{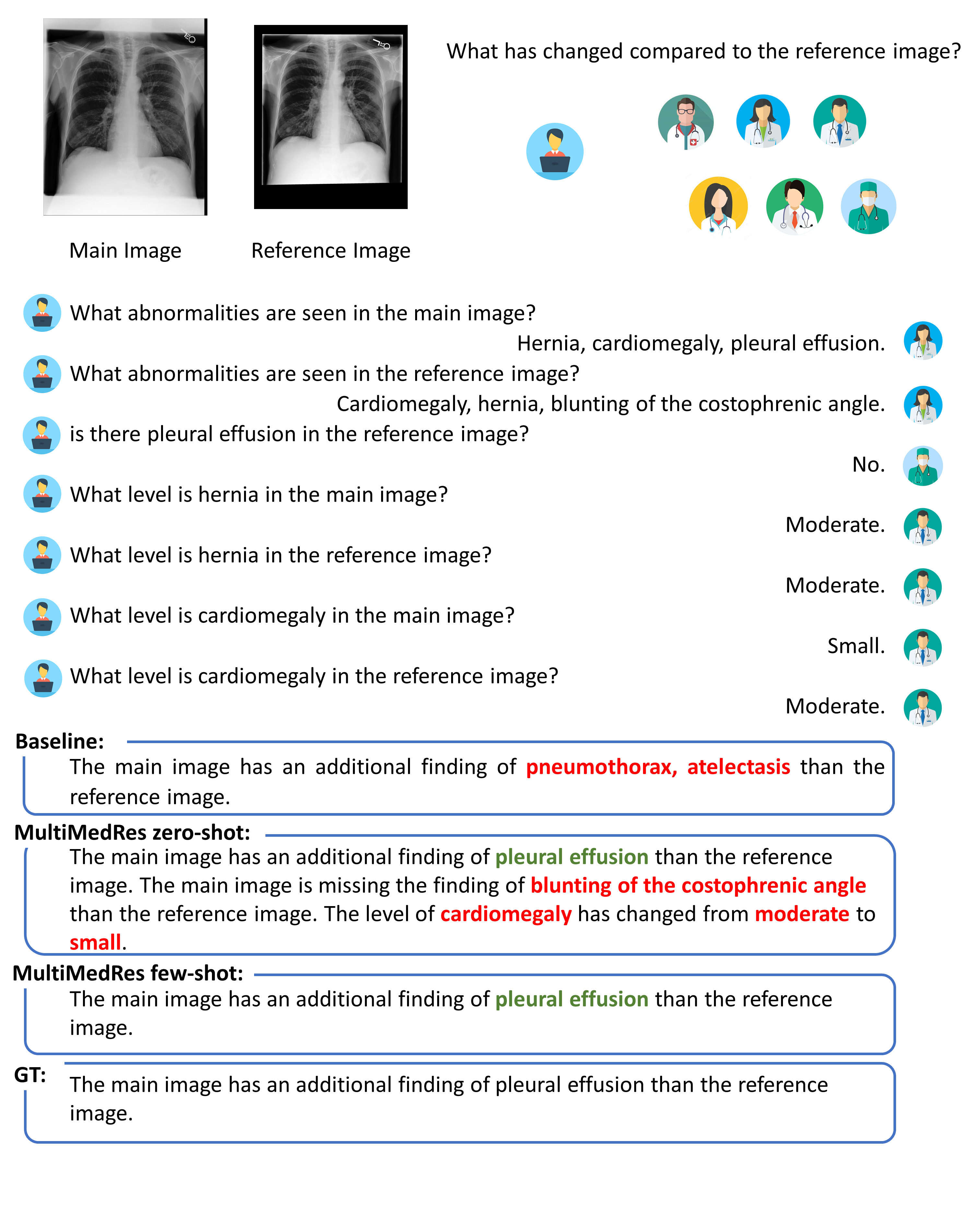}
    \vspace{-7mm}
    \caption{Case study for the commonly-seen question: "What has changed compared to the reference image?". We highlight the same information mentioned in the ground truth answer with the color green, and the redundant or incorrect one with the color red.} 
    \vspace{-6mm}
    \label{fig: 2} 
\end{figure}

\begin{table*}[!htbp]
    \centering
    \scriptsize
    \vspace{-2mm}
    \begin{tabular}{lcccccccc}
        \toprule
        \multirow{2}{*}{\textbf{Gender}} & \multicolumn{4}{c}{Female}         & \multicolumn{4}{c}{Male}           \\
                                         \cmidrule(r){2-5}                    \cmidrule(r){6-9}
                                         & Bleu-4 & METEOR & ROUGE\_L & CIDEr & Bleu-4 & METEOR & ROUGE\_L & CIDEr \\ \midrule
        EKAID                            & 0.398  & 0.342  & 0.556    & 0.805 & 0.417  & 0.357  & 0.583    & 0.860 \\
        MultiMedRes (Zero-shot)             & 0.414  & \textbf{0.358}  & 0.579    & 0.830 & 0.416  & 0.355  & 0.595    & 0.858 \\
        MultiMedRes (Few-shot)               & \textbf{0.454}  & 0.356  & \textbf{0.596}    & \textbf{1.261} & \textbf{0.457}  & \textbf{0.364}  & \textbf{0.608}    & \textbf{1.141} \\
        \bottomrule
    \end{tabular}
    \vspace{-4mm}
    \bigskip
    \setlength{\tabcolsep}{2pt}
    \begin{tabular}{lcccccccccccc}
        \toprule
        \multirow{2}{*}{\textbf{Age}}    & \multicolumn{4}{c}{Age\textless{}55} & \multicolumn{4}{c}{55\textless{}=Age\textless{}70} & \multicolumn{4}{c}{70\textless{}=Age} \\
                                         \cmidrule(r){2-5}                      \cmidrule(r){6-9}                                    \cmidrule(r){10-13}
                                         & Bleu-4  & METEOR  & ROUGE\_L & CIDEr & Bleu-4     & METEOR     & ROUGE\_L     & CIDEr     & Bleu-4  & METEOR  & ROUGE\_L  & CIDEr \\ \midrule
        EKAID                            & 0.364   & 0.350   & 0.537    & 0.887 & 0.426      & 0.352      & 0.585        & 0.849     & 0.445   & 0.348   & 0.597     & 0.743    \\
        MultiMedRes (Zero-shot)             & 0.406   & 0.356   & \textbf{0.597}    & 0.968     & 0.418      & 0.360      & 0.581        & 0.847     & 0.424   & 0.356   & 0.575     & 0.676 \\
        MultiMedRes (Few-shot)               & \textbf{0.432}   & \textbf{0.358}   & 0.586    & \textbf{1.472} & \textbf{0.467}      & \textbf{0.362}      & \textbf{0.609}        & \textbf{1.096}     & \textbf{0.471}   & \textbf{0.362}   & \textbf{0.619}     & \textbf{0.915} \\
        \bottomrule
    \end{tabular}
    \vspace{-2mm}
    \caption{Comparison of baseline against MultiMedRes for two different kinds of split. } 
    \vspace{-5mm}
    \label{tab: bias}
\end{table*}

We present two case studies in \autoref{fig: 2} and \autoref{fig: 3} (see \autoref{sec: rare case study}) to more effectively illustrate our proposed framwork. In addressing the common query, 'What has changed compared to the reference image?', as depicted in \autoref{fig: 2}, the learner agent initially asks about image abnormalities from a global perspective, responses to which are provided by the abnormality detection specialist. Upon identifying the abnormalities in both images, the learner agent then consults the abnormality level specialist to examine the severity of recurring abnormalities, such as \textit{hernia} and \textit{cardiomegaly} in this instance, discovering a change in the level of \textit{cardiomegaly}. Finally, having gathered sufficient information, the agent concludes its inquiries and formulates a comprehensive response incorporating all discussed information. However, this approach may not always yield the most accurate results. For instance, the specialists' responses might occasionally be inaccurate or highlight irrelevant details, as shown in \autoref{fig: 2}. The change in \textit{cardiomegaly} severity from small to moderate may be less critical than the new finding of \textit{pleural effusion}, and the noted \textit{blunting of the costophrenic angle} might be a false positive. Thus, including all these details, as the learner agent does, could lead to a lower evaluation score. Conversely, the supervised few-shot prediction model can potentially rectify these inaccuracies with direct access to the images. This improvement is contingent on access to the chatlog, which enables the model to efficiently identify the correct information through the conversation, as opposed to extracting it by itself. With the chatlog, the search scope for the supervised model is significantly narrowed, yielding enhanced performance with a reduced reliance on training data.

It is worth noticing that the LLM learner agent can also rectify errors made by the specialists. In our observations, the learner agent occasionally poses what appears to be a redundant question, such as verifying the presence of a disease previously indicated by the abnormality specialist, or mentioning an abnormality not identified earlier in the discussion. Interestingly, these additional queries may lead the specialists to reconsider their previous statement, potentially leading to a more accurate final answer. We hypothesize that such questions derive from the extensive knowledge base of the LLMs. For instance, if the LLM rarely encounters instances of \textit{blunting of the costophrenic angle} co-occurring with \textit{cardiomegaly} in prior radiology reports, it might seek verification from the specialists, thereby potentially rectifying any inaccuracies. Indeed, addressing a simple yes-or-no question about a specific abnormality is significantly more straightforward than identifying all abnormalities, and it is evident that our specialists respond more accurately to these direct inquiries, as evidenced in \autoref{tab: VQA results}. Additionally, for less common inquiries such as 'What has changed in the right lung?' shown in \autoref{fig: 3}, the LLM learner agent skillfully tailors its questions to acquire more targeted information. In these scenarios, the zero-shot answer from the learner agent is often more relevant, since the supervised few-shot answer may default to more general response patterns due to limited exposure to such uncommon cases. Given this analysis, the LLM learner agent is undoubtedly an integral component of our proposed framework. And, more importantly, our MultiMedRes is equipped to effectively harness its reasoning abilities and extensive knowledge base to tackle specific domain tasks.

\subsection{Bias Evaluation}
\label{sec: bias}
In this section, we assess the DVQA performance of our approach compared to the previous state-of-the-art, utilizing more fine-grained datasets. Specifically, we further segregate our test set into two subsets based on gender and age. To ensure a relatively even distribution of data and adhere to medical intuition, we categorize ages into three groups: $Age < 55 (29.1\%)$, $55 \leq Age < 70 (33.9\%)$, and $Age \geq 70 (37.0\%)$. The results are presented in Table~\ref{tab: bias}. As observed, the prior method, EKAID, exhibits substantial bias when encountering different age and gender groups, achieving significantly better performance towards elder patients and male patients. Conversely, both the zero-shot predictions made by LLMs and the few-shot predictions made by EKAID with augmented training data effectively reduce such bias. This demonstrates the generalization capability and effectiveness of our method across diverse datasets and patient groups.

%% file: 05conclusion.tex
\section{Conclusion}

In this study, we explore the demanding yet intricate setting of the medical image difference VQA problem. 
To navigate the complexities of identifying differences in medical images in response to specific questions, we introduce the multimodal medical collaborative reasoning framework, MultiMedRes, leveraging the superior reasoning capabilities of text-based LLMs. We first train multiple domain-specific expert models to address questions related to single images. Then, we incorporate the LLM as a learner agent to iteratively generate questions for the expert models, aiming to gather the necessary information in order to address the difference questions. Our extensive empirical analysis confirms the effectiveness of our proposed approach, which consistently improves upon the performance of baseline models under various metrics.

%% file: 06appendix.tex
\section{Implementation Details}

\subsection{Statistics of the DQA Dataset}
\label{sec: statistics}
We examined our proposed technique on the public DVQA dataset, MIMIC-Diff-VQA\cite{10.1145/358}, the statistics of which is presented in \autoref{tab: static}. We adhered to the official splits, with the train/validation/test split being 8/2/2. To ensure the integrity of the evaluation process, the images in different sets (train, validation, and test) do not overlap, effectively preventing any leakage of test images into the training dataset of the domain experts. Additionally, for each question type, we provide two examples. These examples also serve as context-learning material for the LLMs.
\begin{table*}[!htbp]
    \footnotesize
    \centering
    \begin{tabular}{cccc}
    \toprule
    Question Type                & Example                                                                                                                                                        & \# QA pairs     &\# answer candidates        \\ \hline
    \multirow{2}{*}{Abnormality} & \multirow{2}{*}{\begin{tabular}[c]{@{}c@{}}what abnormalities are seen in this image?\end{tabular}}                                                            & \multirow{2}{*}{59,435} & \multirow{2}{*}{8316} \\
                             &                                                                                                                                                      &                         &                \\ \hline

    \multirow{2}{*}{Abnormality*} & \multirow{2}{*}{\begin{tabular}[c]{@{}c@{}}what abnormalities are seen in the upper lungs?\end{tabular}}                                                            & \multirow{2}{*}{85,986} & \multirow{2}{*}{25} \\
                             &                                                                                                                                                      &                         &                \\ \hline
    Presence                     & \begin{tabular}[c]{@{}c@{}}is there evidence of atelectasis in this image?\\ Is there edema?\end{tabular}                                                      & 155,726          & 2        \\ \hline
    View                         & \begin{tabular}[c]{@{}c@{}}which view is this image taken?\\ is this AP view?\end{tabular}                                                                     & 56,265           & 4        \\ \hline
    Location                     & \begin{tabular}[c]{@{}c@{}}where in the image is the pleural effusion located?\\ is the atelectasis located on the left side or right side?\end{tabular}       & 84,193           & 735        \\ \hline
    Type                         & \begin{tabular}[c]{@{}c@{}}what type is the opacity?\\ what type is the atelectasis?\end{tabular}                                                              & 27,478           & 87        \\ \hline
    Level                        & \begin{tabular}[c]{@{}c@{}}what level is the cardiomegaly?\\ what level is the pneumothorax?\end{tabular}                                                      & 67,296           & 112        \\ \hline
    \multirow{2}{*}{All} & \multirow{2}{*}{\begin{tabular}[c]{@{}c@{}}-\end{tabular}}                                                                                                             & \multirow{2}{*}{536,379} & \multirow{2}{*}{9,256} \\
                             &                                                                                                                                                      &                         &                \\ \hline
    Difference                   & \begin{tabular}[c]{@{}c@{}}what has changed compared to the reference image?\\ what has changed in the right lung area?\end{tabular}                           & 164,324        &-          \\ 
    \bottomrule
    \end{tabular}
    \caption{Statistics of the MIMIC-CXR-VQA dataset. \textit{Abnormality*} stands for the abnormality questions excluding the question "what abnormalities are seen in this image?". \textit{All} stands for the questions excluding the difference type of questions. Note that the examples shown here are the learning context we provide to the learner LLM models.}
    \label{tab: static}
\end{table*}

\subsection{Implementation Details}
\label{sec: implementation}
We incorporate gpt-3.5-turbo-0125, gpt-4-1106-preview, and LLaMa2-70B-chat as the learner agents. To enhance the reproducibility and stability of their performance, we set the temperature parameter at 0.2 for both LLMs. In terms of specialist agents, we integrated the MMQ \cite{aioz_mmq_miccai21} as the domain expert VQA model and a 121-layer DenseNet \cite{Huang2016DenselyCC} as the multi-label prediction model for abnormality detection. 
Regarding the training hyperparameters, we adhered to the configurations specified in the original studies and retained the default settings. We train expert models and perform evaluation on an NVIDIA Tesla P100 GPU with 16 GB memory.

\subsection{Baselines}
\label{sec: baseline}
\begin{itemize}
    \item \textbf{MMQ} MMQ\cite{aioz_mmq_miccai21} is a recently proposed VQA model designed for medical images, which adopts Model Agnostic Meta-Learning (MAML) through pretraining multiple meta-models on natural images and finetuning on the medical images.
    \item \textbf{EKAID} EKAID\cite{10.1145/358} aligns the high dimensional feature of different x-ray images through an expert knowledge graph. The model is designed to adaptively choose either to focus on the subtractive difference feature or the main image feature by utilizing the attention mechanism.
    \item \textbf{UIO} UNIFIED-IO\cite{lu2022unifiedio} (UIO) is a unified vision model designed to handle a broad range of vision-language tasks, including VQA, by standardizing diverse inputs and outputs into a sequence of tokens, which enables the model to be trained on over $90$ different datasets using a unified transformer-based architecture.
    \item \textbf{MiniGPT-V2} MiniGPT-v2 \cite{chen2023minigptv2} is a unified large language model designed to efficiently handle a variety of vision-language tasks based on Llama2. By using unique identifiers for different vision-language tasks during training, the model achieves strong performance on multiple benchmarks.
    \item \textbf{LLaVa} LLaVA \cite{liu2023llava}, representing a popular practice for training multi-modal large language models, combines a vision encoder with a language model, which utilizes GPT-4 to generate language-image instruction data for improving the zero-shot capabilities. We incorporate LLava-v1.5 as a baseline in this work.
\end{itemize}

\begin{table}[!htbp]
    \centering
    \begin{tabular}{c|ccc}
    \toprule
    Models          & Open & Close & All  \\ \hline
    MMQ             & 11.5 & 10.8  & 11.5 \\
    EKAID           & 26.4 & 79.9  & 52.5 \\
    UIO             & 0.04 & 53.6  & 35.4 \\
    MiniGPT-v2      & 15.5 & 57.3  & 45.2 \\
    LLaVa           & 13.1 & 55.8  & 42.6 \\
    Ours   & \textbf{54.3} & \textbf{84.8}  & \textbf{70.6} \\
    \bottomrule
    \end{tabular}
    \caption{Comparative performance of various models on questions concerning single images.} 
    \label{tab: VQA results}
\end{table}

\section{Rare Case Study}
\label{sec: rare case study}

\begin{figure}[h]
    \centering
    \includegraphics[width=0.48\textwidth]{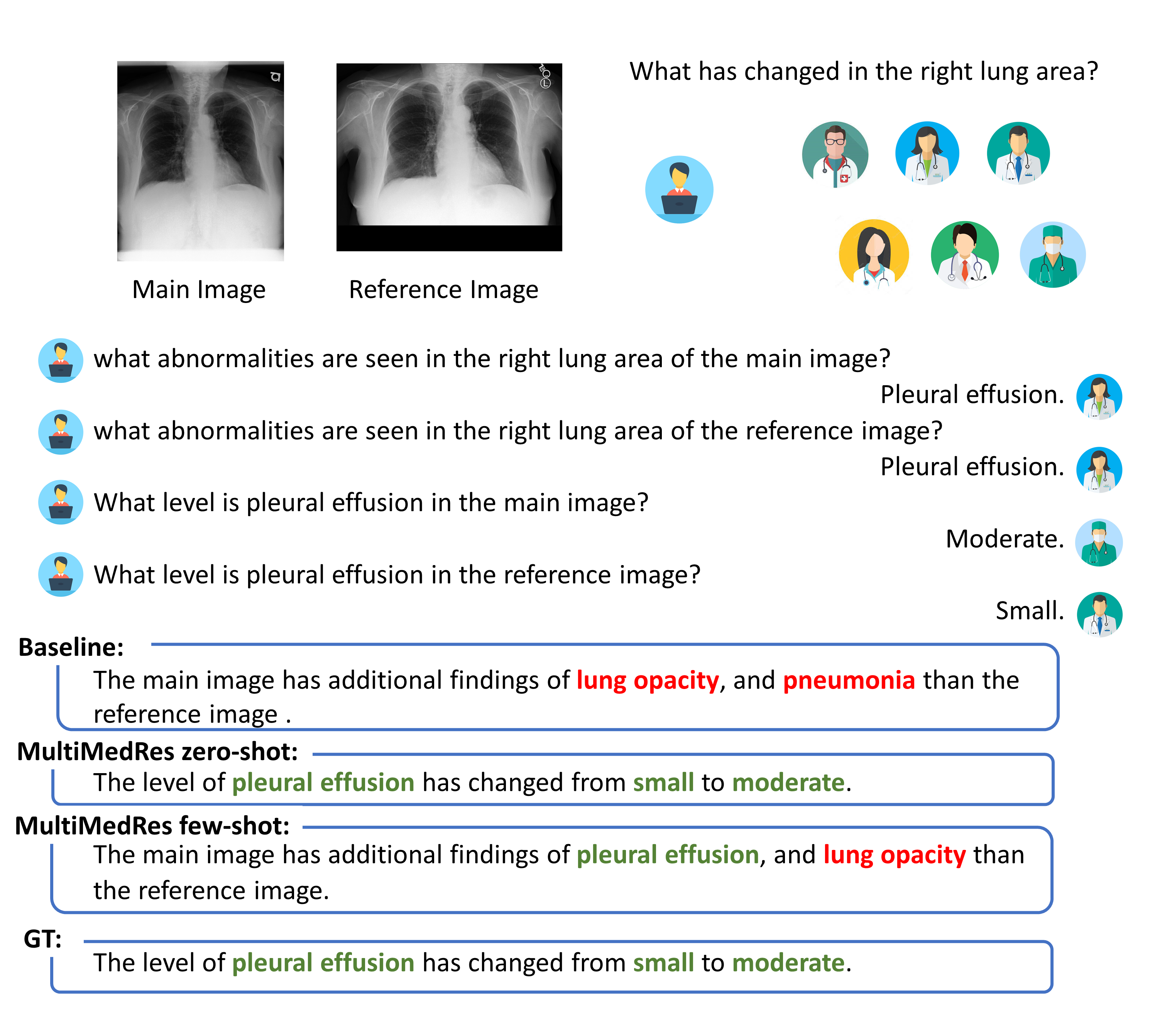}
    \caption{Case study for the rare question: "What has changed in the right lung?". We highlight the same information mentioned in the
ground truth answer with the color green, and the redundant or incorrect one with the color red.} 
    \label{fig: 3} 
\end{figure}

We present a case study focusing on a relatively rare question: "What has changed in the right lung?" in \autoref{fig: 3}. This example conversation demonstrates that the LLM learner agent adapts its generated questions to gather more specific information effectively. We visually differentiate the correspondence between the model-generated responses and the ground truth. Information aligning with the ground truth answer is highlighted in green, whereas information that is either redundant or incorrect is marked in red.

\section{VQA Performance}
\label{sec: VQA performance}

We present the performance of various VQA models on single-image-related questions in \autoref{tab: VQA results}. It can be observed that employing a divide-and-conquer strategy significantly improves performance, thereby enabling our MultiMedRes to acquire accurate information throughout the communication process.

\section{Prompts}
\label{sec: prompt}
\begin{figure*}[h]
    \centering
    \includegraphics[width=\textwidth]{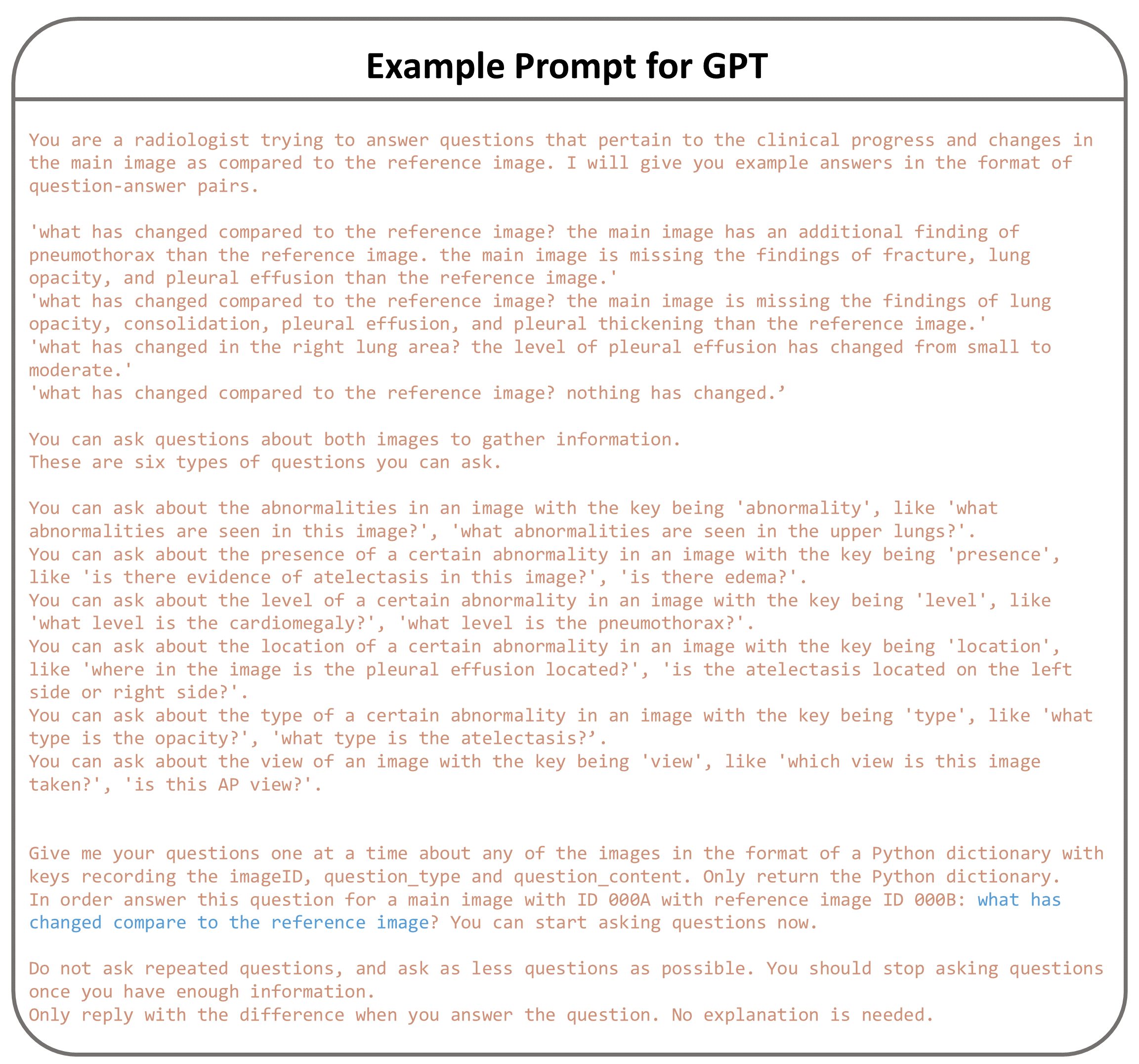}
    \caption{An example of the prompt for GPT.} 
    \label{fig: prompt1} 
\end{figure*}

\begin{figure*}[h]
    \centering
    \includegraphics[width=\textwidth]{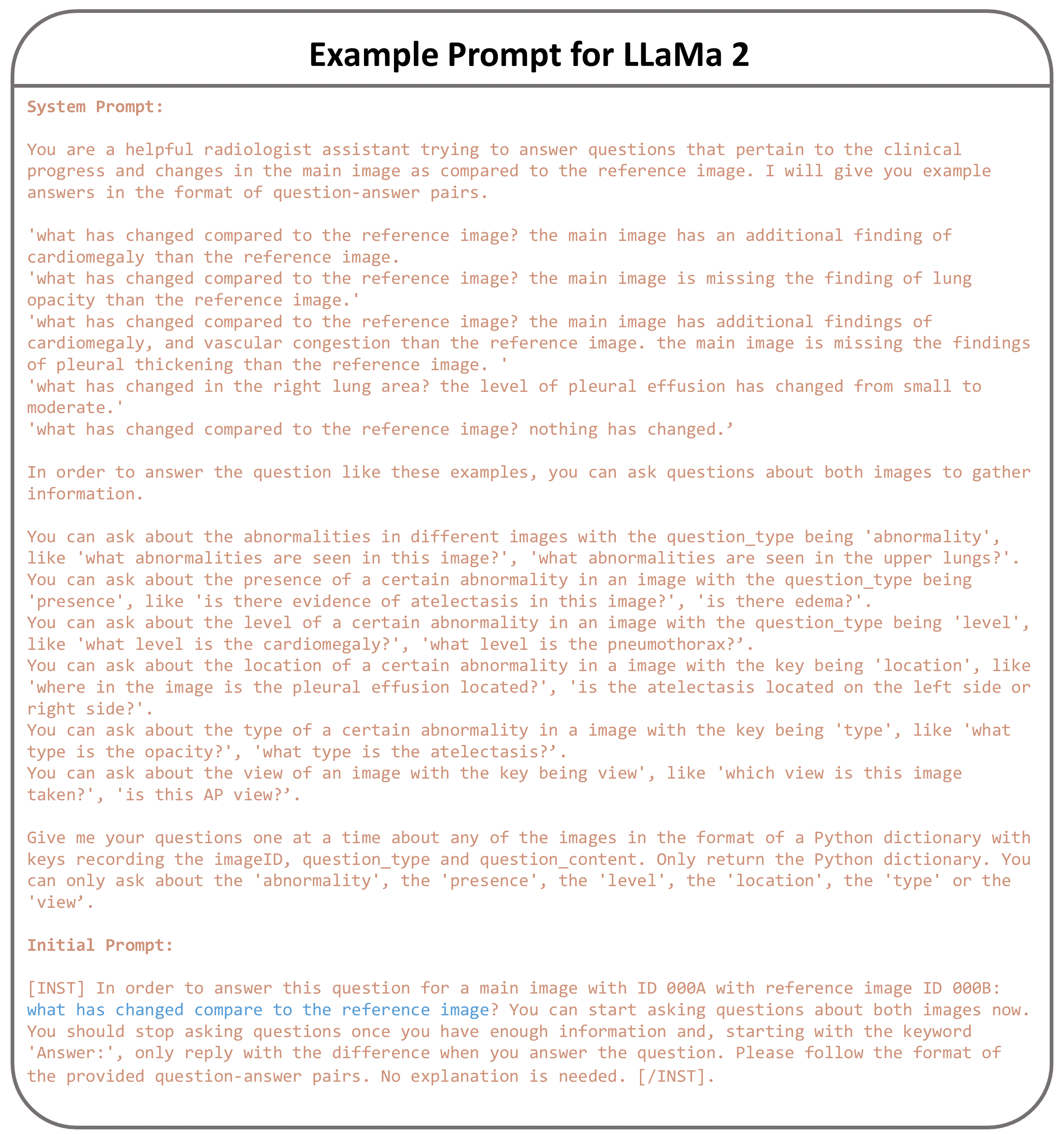}
    \caption{An example of the prompt for LLaMa.} 
    \label{fig: prompt2} 
\end{figure*}

We exhibit two example prompts for GPT and LLaMa 2 in \autoref{fig: prompt1} and \autoref{fig: prompt2} respectively. We change the difference questions (highlighted in blue) accordingly at runtime.

\section{Human Evaluation}
We have conducted a human evaluation experiment to demonstrate the efficacy of our proposed framework. Specifically, we invite two professional physicians to evaluate the perceptual quality of $100$ randomly selected difference questions and the answers generated by EKAID (fully-supervised model), Med-Flamingo (fine-tuned visual LLM) and our MultiMedAgent (Zero-shot). To ensure objectivity, the evaluators were blinded to the origin of the answers and were asked to identify the response most closely aligning with the Ground Truth (GT) answer, or to declare a tie if appropriate. The results, as detailed in \autoref{tab: human}, reveal that MultiMedRes significantly outperforms the baseline models in a clinical setting, as evidenced by its highest pick-up percentages. This finding affirms that the integration of LLM agents with domain-specific expertise not only leverages the strengths of both approaches but also substantially enhances the system’s reliability and clinical applicability.

\begin{table*}[h]
    \centering
    \begin{tabular}{|c|c|c|c|}
    \hline
    EKAID wins & Med-Flamingo wins & MultiMedRes(Zero-shot) Wins & Tie \\ \hline
    21         & 12                & \textbf{48}                 & 19  \\ \hline
    \end{tabular}
    \caption{Human Evaluation.} 
    \label{tab: human}
\end{table*}

\section{Ablation Study}
We have conducted an ablation study to evaluate the impact of the divide-and-conquer strategy and the integration of an abnormality detection specialist within our framework. The results underscore that both the divide-and-conquer approach and the inclusion of the abnormality detection specialist significantly contribute to the enhanced performance of our framework.

\begin{table*}[h]
    \centering
    \begin{tabular}{c|c|cccc}
    \toprule
    Learner Agent                    & Model Variation           & Bleu-4 & METEOR & ROUGE\_L & CIDEr \\ \midrule
    \multirow{3}{*}{LLaMa2-70b-chat} & w/o divide-and-conquer    & 0.299  & 0.321  & 0.495    & 0.412 \\ 
                                     & w/o abnormality detection & 0.320  & 0.356  & 0.532    & 0.456 \\ 
                                     & -                         & 0.345  & 0.373  & 0.554    & 0.483 \\ \midrule
    \multirow{3}{*}{ChatGPT-4-Turbo} & w/o divide-and-conquer    & 0.331  & 0.313  & 0.487    & 0.637 \\ 
                                     & w/o abnormality detection & 0.408  & 0.351  & 0.564    & 0.806 \\ 
                                     & -                         & 0.418  & 0.357  & 0.586    & 0.843 \\ \bottomrule
    \end{tabular}
    \caption{Abaltion Study.} 
    \label{tab: ablation}
\end{table*}